%% file: root.tex
\documentclass[letterpaper, 10 pt, conference]{ieeeconf}  

\IEEEoverridecommandlockouts                              

\usepackage{graphicx} 
\usepackage{amsmath} 
\usepackage{amssymb}  
\usepackage{bm}
\usepackage[ruled]{algorithm2e}
\usepackage{algpseudocode}
\usepackage{textcomp}
\usepackage{lipsum}
\usepackage[table]{xcolor}
\usepackage{array,colortbl}
\usepackage[noadjust]{cite}

\usepackage{xcolor}

\newcommand\etal{\textit{et al. }}
\usepackage{array}
\newcolumntype{P}[1]{>{\centering\arraybackslash}p{#1}}

\DeclareMathOperator*{\argmax}{arg\,max}

\title{\LARGE \bf
FingerSLAM: Closed-loop Unknown Object Localization and Reconstruction from Visuo-tactile Feedback
}

\author{Jialiang Zhao$^{1,2}$, Maria Bauza$^{2}$, and Edward H. Adelson$^{1}$
\thanks{$^{1}$Computer Science and Artificial Intelligence Lab, Massachusetts Institute of Technology
    {\tt\small \{alanzhao, adelson\}@csail.mit.edu}}
\thanks{$^{2}$Mechanical Engineering, Massachusetts Institute of Technology
        {\tt\small bauza@mit.edu}}%
}

\begin{document}

\maketitle
\thispagestyle{empty}
\pagestyle{empty}

\input{body/0_abstract.tex}

\input{body/1_introduction.tex}

\input{body/2_related_works.tex}

\input{body/3_statement.tex}

\input{body/4_method.tex}

\input{body/5_results.tex}

\input{body/6_conclusions.tex}





\section*{ACKNOWLEDGMENT}

Toyota Research Institute provided funds to support this work.
The authors thank Dr. Shaoxiong Wang for providing support for the GelSight sensor, Sandra Q. Liu for helping the authors to make suitable gels, and Branden Romero for the support in setting up the motion capture system. 

\bibliographystyle{IEEEtran}
\bibliography{refs}

\end{document}

%% file: body/0_abstract.tex
\begin{abstract}

In this paper, we address the problem of using visuo-tactile feedback for 6-DoF localization and 3D reconstruction of unknown in-hand objects.
We propose FingerSLAM, a closed-loop factor graph-based pose estimator that combines local tactile sensing at finger-tip and global vision sensing from a wrist-mount camera.
FingerSLAM is constructed with two constituent pose estimators: a multi-pass refined tactile-based pose estimator that captures movements from detailed local textures, and a single-pass vision-based pose estimator that predicts from a global view of the object.
We also design a loop closure mechanism that actively matches current vision and tactile images to previously stored key-frames to reduce accumulated error.
FingerSLAM incorporates the two sensing modalities of tactile and vision, as well as the loop closure mechanism with a factor graph-based optimization framework.
Such a framework produces an optimized pose estimation solution that is more accurate than the standalone estimators.
The estimated poses are then used to reconstruct the shape of the unknown object incrementally by stitching the local point clouds recovered from tactile images.
We train our system on real-world data collected with 20 objects.
We demonstrate reliable visuo-tactile pose estimation and shape reconstruction through quantitative and qualitative real-world evaluations on 6 objects that are unseen during training.

\end{abstract}

%% file: body/1_introduction.tex
\section{Introduction}


Understanding an in-hand object's pose and shape is a fundamental component to many robotic manipulation tasks, such as grasping \cite{ferrari1992planning}, assembly \cite{zhao2020towards}, and tool using \cite{fang2020learning}. 
While pose estimation and shape reconstruction through vision have been active areas of research in the computer vision domain \cite{wang2019densefusion,xiang2017posecnn,hodan2018bop}, physical interactions through touch provide humans with an effective yet often underestimated way to understand the scene \cite{klatzky1985identifying}. 
With the recent advance in high-resolution tactile sensing \cite{wang2021gelsight,kappassov2015tactile}, researchers finally have the means to obtain detailed contact information such as the shape and force in a real-time manner. 
Enabled by these newly developed sensors, several works focus on scene understanding relying solely on high definition tactile sensing \cite{yuan2015measurement,zhang2018fingervision,bauza2019tactile,bauza2022tac2pose,suresh2021tactile,suresh2022shapemap,sodhi2022patchgraph}.

However, humans use both vision and tactile during shape inference and localization \cite{klatzky1987there}.
Vision provides a global understanding of the scene, while tactile provides a detailed view for the contact geometry.
Both modalities are needed for precise localization and reconstruction.
In this work, we tackle the problem of localizing and reconstructing an unknown in-hand object by combining the two sensing modalities.

To solve this problem, we need to answer the following question: How can we calculate an optimal pose given multiple sensory estimates that do not always agree with each other?
Simultaneous Localization And Mapping (SLAM) is a well-studied problem in the field robotics domain \cite{durrant2006simultaneous}.
Central to SLAM is the task of jointly optimizing an objective, often a multi-dimensional pose, by leveraging multiple sensory measurements and producing a more accurate and tractable solution.
One popular choice of such optimization frameworks is the factor graph, which models different sensory measurements and pose priors as factors in a bipartite graph.
Due to its flexibility and computational efficiency, we chose it as the optimization tool for our visuo-tactile pose estimation and shape reconstruction task.

Our main contributions are:
\begin{itemize}
    \item A factor graph-optimized 6-DoF pose estimator that combines global vision and local tactile tracking.
    \item A loop closure matching process that effectively reduces accumulated errors.
\end{itemize}

We train FingerSLAM with 20 real-world objects, and evaluate it on 6 objects that are unseen during training.
FingerSLAM achieves a pose tracking accuracy of $2.8mm / 2.4deg$, and produces realistic reconstructed point clouds.

%% file: body/2_related_works.tex
\section{Related Works}

\begin{figure*}[!ht]
\centering
\includegraphics[width=\linewidth]{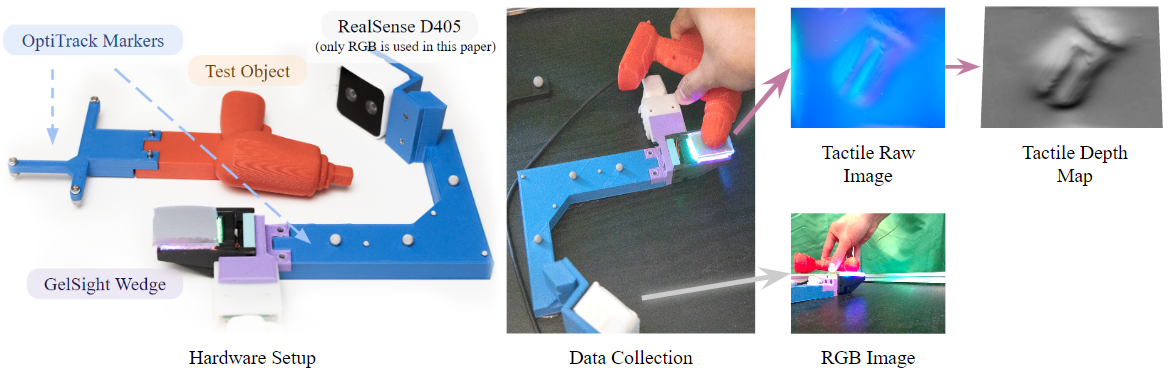}
\caption{\textbf{Hardware Setup.} We use a GelSight Wedge sensor for tactile sensing, an Intel ReslSense D405 camera mounted on the side for RGB vision sensing, and an OptiTrack setup for motion capture. \textbf{Data Collection.} The tactile finger and the camera are fixed to the table at all times. A human operator moves a test object and presses it against the finger. We show sampled tactile and RGB images as well as a reconstructed local tactile depth map on the right.}
\label{datacollection}
\end{figure*}


Researchers of the robotics community have put forward a wide range of tactile sensing solutions.
Sensors working on different sensing principles have been adopted to solve a large set of manipulation tasks.
Among different types of tactile sensors, vision-based ones such as GelSight \cite{yuan2017gelsight} and GelSlim \cite{donlon2018gelslim} stand out for their rich output, ease to use, and affordability.
While we focus on the pose estimation and shape reconstruction task using vision-based tactile sensors, we refer readers to \cite{kappassov2015tactile} for an in-depth review of different types of tactile sensing and their applications.
In this section, we review works on three typical tasks that are most relevant to our solution: slip detection, object property inference, and SLAM.


\textbf{Slip detection and estimation}:
Using a similar sensor to ours, Yuan \etal compared and analyzed a GelSight tactile sensor's images collected at different stages of slip in \cite{yuan2015measurement} and showed this type of sensors' capability in detecting micro scale movements.
Li \etal and Zhang \etal trained recurrent neural networks on tactile images to detect slip between multiple time steps in a manipulation sequence \cite{li2018slip, zhang2018fingervision}.
Built on their binary slip detection model in \cite{li2018slip}, Li further added rotational slip direction prediction in \cite{li2019rotational}.
Calandra \etal improved a grasp planner for the classic robot bin-picking problem by incorporating slip detection and achieved a higher grasp success rate \cite{calandra2017feeling}. 
However, those methods only detect slip without localizing the object after the slip.
In many precision manipulation tasks we are also interested in the amount of the displacement.

\textbf{Object property inference and localization}:
Many works have focused on inferring properties of the in-contact object, such as shape \cite{strub2014using, luo2015tactile, luo2019iclap}, texture \cite{luo2018vitac, yuan2017connecting}, and material \cite{yuan2017connecting, kroemer2011learning, kerr2018material}.
Those learned object properties can be further used for localization.
In order to localize current grasps, Bauza \etal proposed to match new tactile imprints with previously collected tactile imprints \cite{bauza2019tactile}, while Luo \etal learned to match tactile imprints directly to visual images of the whole object \cite{luo2015localizing}.
Assuming known CAD models, Bauza \etal proposed to localize by comparing contact masks generated from tactile images with a large bank of random projections of the CAD model \cite{bauza2022tac2pose}.
To solve the reverse problem, i.e. what a tactile image looks like given an object and a pose, several tactile simulators have been built to automatically generate tactile images given an object's CAD model and a finger pose \cite{si2022taxim, wang2022tacto}.
One major limitation for this category of works is that they all require a known calibrated geometry of the object: a pre-collected tactile map \cite{bauza2019tactile}, a model of the object \cite{bauza2022tac2pose}, or a global image with known geometry \cite{luo2015localizing}.
This requirement can be hard to meet in less constraint environments.

\textbf{Tactile SLAM}:
Recent studies have shown interests in working with unknown objects by leveraging methods from the SLAM problem.
With a focus on 2D shapes, Suresh \etal parameterized shapes as Gaussian Process Implicit Surfaces (GPIS), and learned its parameters from tactile signals collected during pushing \cite{suresh2021tactile}.
Assuming known contact poses, authors of \cite{suresh2022shapemap} first learned a noisy mapping from known surface geometries to corresponding tactile images, then reconstructed an object by combining many noisy local tactile measurements into an optimized global shape using factor graph optimization.
The closest prior work to ours is \cite{sodhi2022patchgraph}, where the authors learned to estimate 6D poses and 3D shapes simultaneously for unknown objects. 
They constructed a pose estimator based on tactile sensing, and a shape reconstruction pipeline that added in new tactile point clouds incrementally on the run.
However, this approach heavily relies on the performance of the tactile pose estimator, which lacks a global understanding of the object and can suffer from repeated patterns or smooth surfaces.
In contrast, our work combines vision and tactile sensing which provides us with both global and local understandings of the scene without requiring any other domain knowledge.
Furthermore, we designed a loop closure mechanism that periodically matches current tactile and vision images to stored key-frames, which significantly reduced accumulated errors.
With this, FingerSLAM is able to produce realistic reconstructions even in long sequences. 

%% file: body/3_statement.tex
\section{Problem Formulation}

We consider the problem of estimating an unknown object's pose while simultaneously reconstructing its shape.
Given as inputs a tactile image collected from the finger and a RGB image collected from a wrist-mounted camera, FingerSLAM estimates the object's 6-DoF pose $P_t \in SE(3)$ and reconstructs its shape as a point cloud $S_t$ incrementally, at each time step $t$.
We show our hardware setup and sample images in Fig. \ref{datacollection}.

Assumptions:
\begin{itemize}
    \item Only one object is present in the same sequence.
    \item The object is rigid.
    \item The object is always in contact with the finger, and it is always inside the wrist-mounted camera's view.
\end{itemize}

The remainder of this paper is organized with our approach for pose estimation and reconstruction in Sec. \ref{sec:approach} and the discussion of performance on evaluated objects in Sec. \ref{sec:eval}.

%% file: body/4_method.tex
\section{Localization and Reconstruction}

\begin{figure*}[ht]
\centering
\includegraphics[width=0.8\linewidth]{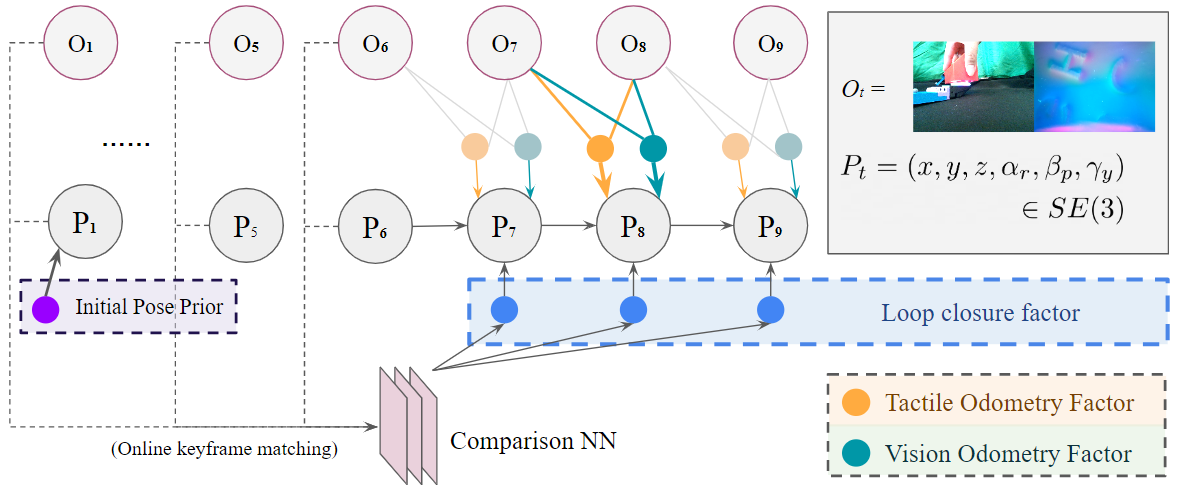}
\caption{Factor graph formulation. $O_t$ denotes the observation (tactile and vision images) at each time step, while $P_t$ denotes the estimated $SE(3)$ pose at each time step. Each factor is denoted as color dots.}
\label{factorgraph}
\end{figure*}

\label{sec:approach}
We begin with pre-processing steps for tactile images, follow with our factor graph formulation of the localization problem, end with our reconstruction approach.

\subsection{Local Surface Reconstruction From Tactile Images}
\label{sec:preprocess}
Local surface depth maps are reconstructed with photometric stereo for the GelSight Wedge \cite{wang2021gelsight} sensor that we used in this project.
On the outside of a GelSight Wedge sensor is a thin layer of elastic silicone gel, which is coated with reflective paint.
Once pressed, the gel conforms to the contact surface, and its inner side is illuminated by light sources placed at three sides of the gel with three different colors.
A color image is taken, and the color value at each pixel is correlated to the gradients at two directions $\nabla_x, \nabla_y$ at that location. 
During post-processing, such images are first unwarped to remove optical distortion.
Then the color image is converted to a gradient map using a pre-calibrated look-up table.
At each pixel, the gradients $\nabla_x, \nabla_y$ are recovered. 
A Poisson solver is used to reconstruct a depth map from this gradient map. 
An example raw tactile image and its corresponding reconstruction can be found in Fig. \ref{datacollection}.


\subsection{Pose Estimation with Factor Graph}
\label{sec:poseestimate}
Following works in the tactile SLAM domain \cite{suresh2021tactile, sodhi2022patchgraph}, we choose factor graph to synthesize pose measurements from multiple learned estimators. 
A factor graph is a bipartite graph, with factors $F$ as constraints and variables $P$ as optimization targets. 
In our case, $P$ is the object's 6-DoF pose at all time steps, and $F$ consists of a prior $F_{pri}$ and three measurements: tactile odometry $F_{tac}$, vision odometry $F_{vis}$, and loop closure $F_{lc}$.
The optimization problem solves for the best $\bm{P}^* = P_{t=0:N}^*$ such that its posterior probability is maximized (MAP). 
Measurement noise for each factor is modeled as a zero-mean Gaussian distribution, with a covariance matrix $\Sigma$.

\begin{align*} 
\bm{P}^* = \argmax_{\bm{P}} & \Biggl[ ||F_{pri}(P_0)||_{\Sigma_p}^2 + \sum_{t=1}^N \Bigl( ||F_{tac}(P_{t-1}, P_t)||_{\Sigma_t}^2+\Bigr.\Biggr.\\
&\Biggl.\Bigl.||F_{vis}(P_{t-1}, P_t)||_{\Sigma_v}^2 + ||F_{lc}(P_t)||_{\Sigma_l}^2\Bigr) \Biggr]
\end{align*}

Our factor graph formulation is illustrated in Fig. \ref{factorgraph}.
We used the GTSAM C++ library \cite{dellaert2012factor} for factor graph optimization.
Next, we discuss details for each factor.

\textbf{Prior}: 
A unary prior is added to the factor graph only at the initial time step of an episode. 
This prior anchors the object's estimated pose sequence at a given initial pose.
Covariance for the noise model $\Sigma_p$ is zero.

\begin{figure}[ht]
\centering
\includegraphics[width=\linewidth]{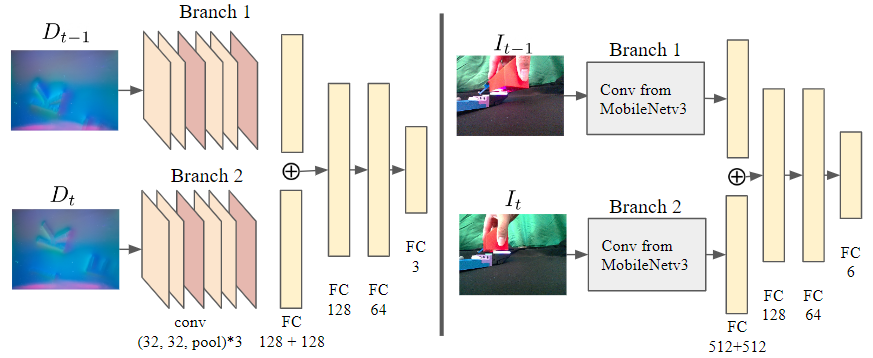}
\caption{Neural network architecture for tactile odometry and vision odometry. (left) The tactile odometry network is composed of two convolutional branch models followed by three fully connected (FC) layers. (right) The vision odometry is constructed similarly to the tactile odometry, with the branch model being replaced with convolutional layers extracted from a pre-trained network, MobileNetv3. In both networks, the two branch models share weights.}
\label{nn}
\end{figure}

\textbf{Tactile odometry with multiple-step refinement}:
Given tactile depth maps collected at two consecutive time steps that sufficiently overlaps with each other, $D_{t-1}$ and $D_t$, it it possible to tell the relative movement of the object $P_{t-1}^t$ between the two time steps.
We learn the mapping $P_{t-1}^t = f_{tacNN}(D_{t-1}, D_t)$ using convolutional neural networks. 
The architecture of $f_{tacNN}$ is illustrated on the left of Fig. \ref{nn}.

\begin{figure}[ht]
\centering
\includegraphics[width=\linewidth]{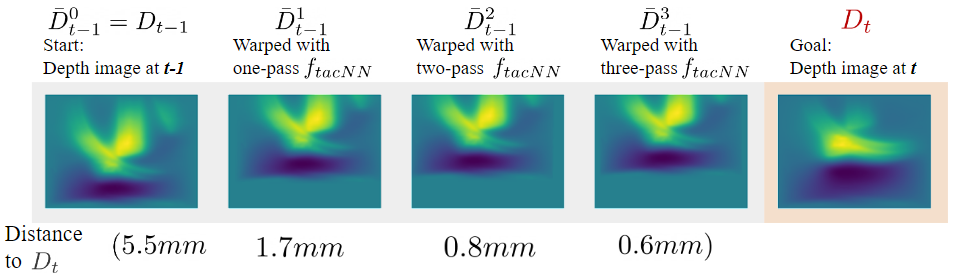}
\caption{A sample sequence of intermediate and goal tactile images from multi-step refined pose estimation. $D_{t-1}$ and $D_t$ are the raw tactile images collected at time step $t-1$ and $t$. Intermediate $\Bar{D}_{t-1}^i$'s are warped versions of $D_{t-1}$ according to the estimated pose at each step. After the last step $i=3$ in the refinement process, the error was reduced to $0.6mm$, compared to the original movement of $5.5mm$. Visually, $\Bar{D}_{t-1}^i$ gets closer to $D_t$ as $i$ increases.}
\label{tactilestage}
\end{figure}

Inspired by \cite{wang2019densefusion}, we propose a multi-pass refinement process which runs $f_{tacNN}$ multiple times until convergence.
This process is detailed in Alg. \ref{alg:tacref}.
A sequence of resulted intermediate $\Bar{D}_{t-1}^i$ as well as $D_t$ are shown in Fig. \ref{tactilestage}.
As $i$ (the number of passes) goes larger, $\Bar{D}_{t-1}^i$ appears closer to $D_t$, which means the estimated pose improves over time.

\begin{algorithm}
\caption{Tactile pose estimator with multi-step refinement}\label{alg:tacref}
\KwData{$D_{t-1}, D_t$}
\KwResult{$P_{t-1}^t$}
$P_{t-1}^t \gets \begin{bmatrix}
    I_{3\times 3} & [0]_{3\times 1}\\
    0 & 1
\end{bmatrix}$\;
$\Bar{D}_{t-1}^0 \gets D_{t-1}$\;
$i \gets 1$\;
$\sigma \gets$ a convergence threshold chosen empirically\;
\While{$i \leq 10$}{
  $\Delta P \gets f_{tacNN}(\Bar{D}_{t-1}^{i-1}, D_t)$\;
  \Comment{Run tacNN to obtain a delta estimate}
  
  $\Bar{D}_{t-1}^i \gets warp(\Bar{D}_{t-1}^{i-1}, \Delta P)$\;
  \Comment{Warp $\Bar{D}_{t-1}^{i-1}$ with the delta estimate}
  
  $P_{t-1}^t \gets \Delta P \cdot P_{t-1}^t$\;
  \Comment{Apply the delta estimate to the result}
  
  \If{$||\Delta P|| < \sigma$}{
    \Comment{Break early if converged}
    
    \Return{$P_{t-1}^t$}\;
  }
  $i \gets i + 1$\;
}
\Return{$P_{t-1}^t$}\;
\end{algorithm}

In experiments we found that the neural network $f_{tacNN}$ learned better when only predicting $SE(2)$ poses, $\{ x, y, \gamma_{yaw} \}$, likely due to the fact that the tactile sensor has a flat surface and a relative thin gel, which means the sensor is only sensitive to tangential movements.
As such, we only predict $\{ x, y, \gamma_{yaw} \}$ with this odometry, and populate $\{ z, \alpha_{roll}, \beta_{pitch} \}$ as all zero, and assign large values to the corresponding cells in the covariance matrix $\Sigma_t$ for the noise model.
Values in other cells of $\Sigma_t$ are chosen empirically.

\textbf{Vision odometry}:
Taken input of RGB images captured from a wrist-mounted camera at two consecutive time steps, $I_{t-1}$ and $I_t$, another CNN $f_{visNN}$ is trained to learn the mapping $P_{t-1}^t = f_{visNN}(I_{t-1}, I_t)$.
The architecture is illustrated on the right of Fig. \ref{nn}.
Unlike $f_{tacNN}$, $f_{visNN}$ predicts the full $SE(3)$ pose because visual images capture global information.
We run it for only one pass at each time step.
Following examples of many works on visual perception, we bootstrap $f_{visNN}$ with feature layers of models that are pre-trained on a larger image dataset.
MobileNetV3 \cite{howard2017mobilenets} is chosen for this task, due to its compact size.
We extract the convolutional layers from the pre-trained model and follow them with four fully connected layers.
The pre-trained layers were frozen at the beginning of training, then they were un-frozen to allow for fine-tuning.
Noise covariance $\Sigma_v$ is chosen empirically.

\textbf{Loop closure}:
Starting from the $3^{rd}$ time step, loop closure is checked between each new time step and all previous time steps that are at least 3 steps earlier, i.e. between the two time steps $t_{i, i>2}$ and $t_{j, j=\forall [1, i-2]}$ we check $LC(i, j) = \{ True, False\}$.
Two neural networks are trained for this process, $G_{tac}(D_{i}, D_{j})$ and $G_{vis}(I_i, I_j)$.
Both output a scalar value between $0$ and $1$ which represents the likelihood that there is a loop closure between $t_i$ and $t_j$.
$G_{tac}(D_{i}, D_{j})$ takes tactile depth maps as inputs and $G_{vis}(I_i, I_j)$ takes RGB images from the wrist-mounted camera as inputs. 
Both neural networks share the same architecture as the odometry networks but with a different output layer, whose size is equal to one.
Then, we let 
\begin{equation}
    \label{eq:lc}
    LC(i, j) = a \times G_{tac}(D_{i}, D_{j}) + b \times G_{vis}(I_i, I_j) \overset{?}{>}  c
\end{equation}
with $a, b$ as weights and $c$ as a threshold.
All three values are chosen empirically.

If any pair of two time steps is considered as a match, we run both the vision odometry and the multi-pass tactile odometry between this pair, and add the measurement to the factor graph.

\subsection{Shape Reconstruction}
\label{sec:reconstruction}
With tactile depth maps $D_t$ and the optimized object's pose $P_t$ at each time step, we reconstruct the object's shape by stitching all $D_i$'s together.
Since the reconstructed shape is not used for any further prediction, the reconstruction is done trivially by scaling $D_t$ to its true size, converting it to a point cloud, anchoring the point cloud to pose $P_t$, and finally reducing the whole point cloud with voxel down-sampling.
Voxel size is chosen empirically.

%% file: body/5_results.tex
\section{Experiments and Evaluation}
\label{sec:eval}

\subsection{Experiment Setup}
We collected real world data $(D_t, I_t, P_t)$'s for both training and evaluation. 
In this section we discuss our data collection setup, with the procedure shown in Fig. \ref{datacollection}.


\textbf{Sensors}:
We used a GelSight Wedge finger for tactile sensing and an Intel RealSense\textsuperscript{TM} D405 camera mounted on the side of the GelSight sensor to mimic the view of a wrist-mounted camera.
Note that even though the D405 camera is a near-range RGB-D camera, we found its depth sensing to be rather noisy in our settings, thus only RGB information was used.

\textbf{Motion Capture}:
We used 4 OptiTrack Flex-3\textsuperscript{TM} cameras to capture the ground truth movement of the object.
We attached a number of reflective markers to the base of the finger and the base of the object.
During experiments, the finger and the RGB camera were always fixed on the desk. 
A human operator held the object and pressed the object against the finger tip at different locations.
Images from both sensors $(D_t, I_t)$ as well as the ground truth pose of the object in the finger's coordinate frame $P_t$ were recorded at each time step.

\begin{figure}[!ht]
\centering
\includegraphics[width=0.75\linewidth]{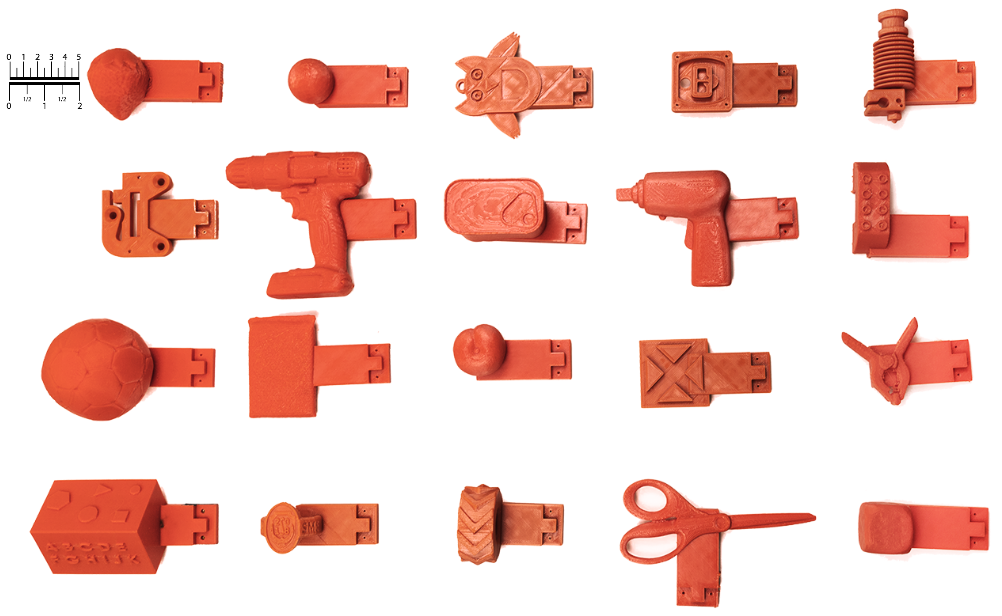}
\caption{Objects used during training. We selected 20 objects from the YCB and the ABC dataset.}
\label{trainobj}
\end{figure}

\textbf{Objects}:
We used 20 3D-printed objects for training, as well as 4 3D-printed and 2 real-world objects for evaluation.
The training objects are shown in Fig. \ref{trainobj}.
Each individual evaluated object will be discussed in Sec. \ref{sec:evalres}.
All objects were selected from the YCB object dataset \cite{calli2015ycb} or the ABC object dataset \cite{Koch_2019_CVPR}.
The two real-world objects are identical glass jars, except one was painted red and the other one was clear.
The two jars were used to test the generalizability of our pose estimator.
We collected about 350 data points on each object, which gave us roughly 7,000 raw training data and 2,500 raw evaluation data.
Because of the difficulty in collecting a larger amount of data in the real world, we augmented the training data in two ways:
\begin{itemize}
    \item For the tactile pose estimator, we artificially translate and rotate the depth images randomly. 
    We first generated multiple random $SE(2)$ poses, $\hat{p}$'s, for each $(P_{t-1}^t, D_{t-1}, D_{t})$ pair, and we applied $\hat{p}$'s to $P_{t-1}^t$ and $D_{t-1}$.
    The generated $(\hat{P}_{t-1}^t, \hat{D}_{t-1}, D_{t})$'s were used as augmented training data.
    \item For both tactile and vision, instead of only using data points collected at consecutive time steps, we also used data points collected at further away time steps as long as the relative displacement between them was smaller than $12mm$.
\end{itemize}
The final size of our training dataset was 40,000 for tactile and 25,000 for vision.
No augmentation was done to the evaluation data.

\subsection{Pose Tracking Performance}
\label{sec:evalres}
In this section two evaluation methods and their results are discussed: 
(1) an ablation study of the pose estimator's per-component performance, and (2) a comparison between the proposed pose estimator's performance and the size of the raw movement.

\textbf{Ablation study}:
Detailed numerical results of the proposed system's per-component performance on each object are shown in Tab. \ref{tab:ablation}. 
The \textit{Raw} column shows the median of the raw movement between two time steps.
This was calculated from the poses collected from the motion capture system.
The rest of the four columns shows the median error obtained from each corresponding pose estimator.

\textit{Tactile v. Vision}:
Comparing the top three objects against the bottom three, we see a clear trend that the tactile estimator performed better when there were unique and fine textures on the object.
We found the tactile estimator to perform the worst on the \textit{Mini Bleach} object, likely due to its mostly smooth surface.
The vision estimator struggled on the two jars, which we believe was attributed to their significantly larger sizes than all the training objects.
Comparing the two jars, tactile performed almost equally, while vision performed worse on the transparent one.
This shows that estimating with tactile has the advantage of generalizability across apparent properties.

\textit{Factor graph optimization}: 
The \textit{Combined} column shows prediction errors of the optimized estimator which combined tactile and vision, without loop closure detection.
For the bottom 4 objects, the \textit{Combined} method achieved a performance in between the performances of tactile standalone and vision standalone.
For the top 2 objects, \textit{Combined} outperformed both standalone methods.
In those cases, the factor graph was able to retrieve good individual channels and compose a 6-DoF prediction that had a higher accuracy.
Moreover, the combined method produced very close accuracy on the two jars, despite the vision estimator's much poorer performance on the transparent jar.
This further confirms that the combined method was able to pick more valuable channels between the estimators.

\textit{Closed-loop factor graph}:
The last column shows the performance of FingerSLAM, the factor graph-optimized estimator with loop closure checking.
We did parameter tuning on the loop closure checking threshold.
The best result was obtained with $a = 0.3, b = 0.7, c = 0.9$ in Eq. \ref{eq:lc}.
Vision was assigned a higher weight for this loop closure process, because tactile suffered more from repeated patterns.
Looking at the \textit{w/ Loop Closure} column, FingerSLAM consistently outperformed the \textit{Combined} method for all objects.
In examinations of per-step errors within an episode, we found that loop closure significantly helped cases when a newly visited spot was close to a previously visited spot.

\begin{table*}[!htb]
      \centering
        \caption{P.E. error per component per object}
        \label{tab:ablation}
        \begin{tabular}{c|c|P{5em}|P{5em}|P{5em}|P{5em}|P{6.9em}}
    \hline
    Name & Picture & Raw\newline mm : deg & Tactile only \newline  mm : deg & Vision only \newline  mm : deg & Combined \newline  mm : deg & w/ Loop Closure \newline  mm : deg \\
        \hline
    Letter Cube 
    & \parbox[c]{10em}{\includegraphics[width=9em]{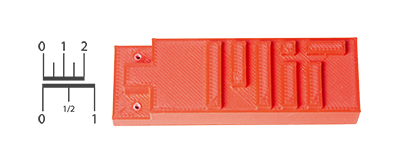}} &
    6.18  : 6.34 & 3.10  : 2.79   & 2.99  : 2.61   & 2.87  : 2.29   & 2.27  : 2.18  \\
    Graphic
    & \parbox[c]{10em}{\includegraphics[width=9em]{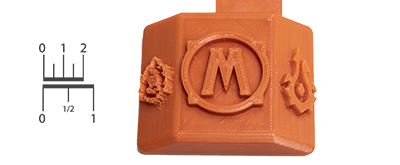}} &
    5.93  : 5.59 & 2.45  : 2.61   & \cellcolor{green!20} 2.86  : 2.11   & \cellcolor{green!20}2.28  : 2.13   & \cellcolor{green!20}2.13  : 2.07  \\
    Rubik's Cube
    & \parbox[c]{10em}{\includegraphics[width=9em]{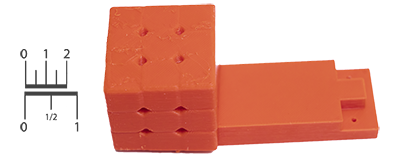}} &
    5.88  : 5.66   & \cellcolor{green!20} 2.33  : 2.80   & 2.97  : 2.37   &  2.50  : 2.43   & 2.30  : 2.41  \\
    Mini Bleach
    & \parbox[c]{10em}{\includegraphics[width=9em]{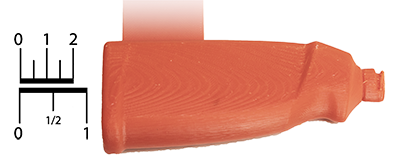}} &
    5.95  : 5.74   & \cellcolor{red!20}4.52  : 3.77   & 3.25  : 2.54   & \cellcolor{red!20}4.06  : 2.91   & \cellcolor{red!20} 3.46  : 2.50  \\
    Painted Glass Jar 
    & \parbox[c]{10em}{\includegraphics[width=9em]{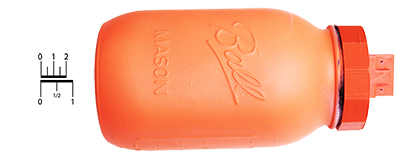}}  & 
    6.40  : 6.21   & 3.50  : 2.70   & 4.12  : 2.94   & 3.63  : 2.63   & 3.19  : 2.82  \\
    Glass Jar 
    & \parbox[c]{10em}{\includegraphics[width=9em]{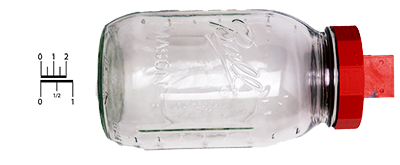}} &
    6.46  : 6.02   & 3.58  : 2.47   & \cellcolor{red!20}4.79  : 4.21   & 3.79  : 2.27   & 3.38  : 2.27  \\
     \hline
  \end{tabular}
\end{table*}

\textbf{Performance v. step size}:
A pose estimator's error hugely depends on the size of the original movement.
We grouped all our evaluation data points in three groups according to the size of their original movement: (a) data pairs that has a movement of less than $5mm$, (b) data pairs of movement between $5mm$ and $10mm$, and (c) data pairs with more than $10mm$ movement.
Detailed per-component performance for each movement group is shown in Tab. \ref{tab:stepsize}.
When step sizes are small (the $<5mm$ group), tactile sensing provides a more accurate estimation compared to vision, which we believe attributed to the fact that tactile has a higher resolution for local movements.
The situation reversed when the step size was large (the $>10mm$ group), likely due to the increased difficulty in finding correlations from two tactile images when the overlapping area is small.
Averaging all our evaluation data (the $all$ row), the tactile estimator performed better than the vision estimator, and our proposed closed-loop combined method achieved the best overall pose estimation result.

\vspace{-1em}

\begin{table}[!htb]
\renewcommand*{\arraystretch}{1.4}
      \caption{P.E. error at each raw movement size group}
      \label{tab:stepsize}
      \centering
        \begin{tabular}{c|c|c|c|c|c}
    \hline
Raw & \scriptsize Raw Median \par& Tactile & Vision & Combined & w/ LC \\
Group& mm:deg& mm:deg& mm:deg& mm:deg& mm:deg\\
\hline
\scriptsize$<$5mm \par& 3.1 : 3.8 & 1.0 : 2.3  & 2.6 : 2.7  & 1.8 : 2.3  & 1.4 : 2.3  \\
\scriptsize5-10mm \par& 7.1 : 5.7  & 3.3 : 2.9  & 3.9 : 3.0  & 3.8 : 2.8  & 2.8 : 2.8  \\
\scriptsize$>$10mm \par& 11.7 : 5.9  & 6.8 : 3.1  & 6.1 : 3.2  & 6.3 : 3.1  & 6.0 : 3.1  \\
\hline
all & 6.7 : 5.6  & 3.0 : 2.7  & 3.5 : 2.9  & 3.2 : 2.4  & 2.8 : 2.4  \\
 \hline
  \end{tabular}
\end{table}

\vspace{-1em}

\subsection{Reconstruction Performance}
\begin{figure}[ht]
\includegraphics[width=\linewidth]{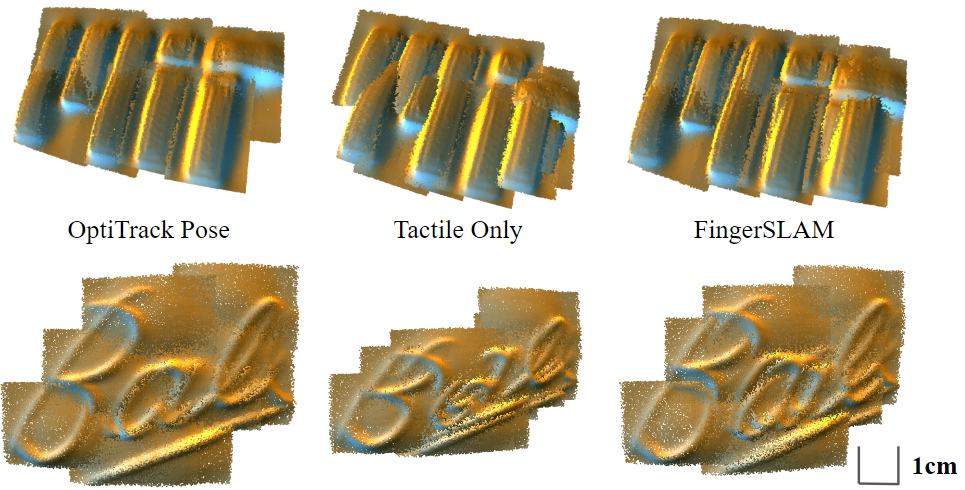}
\caption{Shaped reconstructed from 21 local tactile patches, with poses produced by the motion capture system (left), the tactile pose estimator only (middle), and the proposed estimator (right).}
\label{reconstruction}
\centering
\end{figure}

Qualitative reconstruction results of two objects are shown in Fig. \ref{reconstruction}.
Each sequence used for reconstruction consists of 21 time steps, with every 7 time steps as an episode.
After each episode, the factor graph was reset and a new pose from the motion capture system was used as the initial pose prior for the next episode.
Comparing shapes reconstructed from the tactile pose estimator only (middle column) and the proposed FingerSLAM pose estimator (right column), FingerSLAM wins clearly with much less mis-alignments.
Comparing shapes reconstructed from OptiTrack poses (left column) and FingerSLAM (right column), we see that the two columns share similar overall shape.
The right column shapes are easily recognizable, and appear consistent with the left column despite small mis-alignments in certain areas.

%% file: body/6_conclusions.tex
\section{CONCLUSIONS}

In this paper, we present an approach to localize and reconstruct an unknown shape through vision and tactile feedback.
A factor graph-based framework is proposed to jointly optimize all measurement units, and we close the loop by actively checking for loop closures.
Such a system has the advantage of utilizing both global features and local textures of an object for pose tracking and reconstruction.
Quantitative and qualitative evaluations show that our proposed method is able to achieve high tracking accuracy and produce high-fidelity reconstructed shapes on novel objects.

In the future, this approach can be potentially improved by utilizing force vectors at the contact surface, which encapsulate unique higher order information of the movement.
Furthermore, following works on neural probabilistic inference \cite{zhao2019towards}, instead of using fixed covariance matrices for each factor, having the networks predict both the most likely pose and its variance can potentially help the factor graph make better decisions.